# CricRAG: Retrieval Augmented Vision-Language Models for Personalized Cricket Coaching

**Agamdeep Singh**[1], **Sujit PB**[1], **Mayank Vatsa**[2]
[1]IISER Bhopal, [2]IIT Jodhpur
{agamdeep20, sujit}@iiserb.ac.in, mvatsa@iitj.ac.in

## Abstract

Vision-Language Models (VLMs) offer promising capabilities for automated sports coaching but face a fundamental limitation: they implicitly compare against professional standards, making their feedback impractical for developing players. We present CricRAG, a retrieval-augmented framework that aligns VLMs with skill-appropriate benchmarks for personalized cricket coaching. Our key insight is that by retrieving similar-but-better techniques as reference points, we can guide VLMs to provide developmentally appropriate feedback that mirrors human coaching practices. We contribute: (1) a labelled dataset of 288 cricket technique videos spanning multiple skill levels, (2) an efficient motion retrieval pipeline using contrastive learning that achieves 78% top-3 retrieval accuracy, (3) a frame sampling technique that reduces inference costs, and (4) a retrieval-augmented approach that significantly improves feedback alignment with coaching principles, achieving up to 94% agreement with professional assessments compared to 67% without retrieval context.

## Introduction

Sports coaching fundamentally shapes athletic development, yet access remains heavily constrained by geography and economics. This disparity is particularly acute in cricket-passionate nations like India, where aspiring players lack professional guidance. While Vision-Language Models (VLMs) show potential for democratizing coaching through automated feedback, their direct application faces a critical challenge: they evaluate technique against an implicit professional standard embedded in their pretraining data.

This bias towards professional execution creates a fundamental mismatch with effective coaching practices. Human coaches employ curriculum-based approaches, providing feedback calibrated to a player's development stage and suggesting incremental improvements. VLMs, in contrast, often generate feedback that is both too critical and impractically ambitious for developing players. This limitation stems from two factors: (1) VLMs lack explicit benchmarks for different skill levels, and (2) they struggle with in-context learning for visual tasks (Jiang et al. 2024), making it difficult to align their feedback with coaching principles.

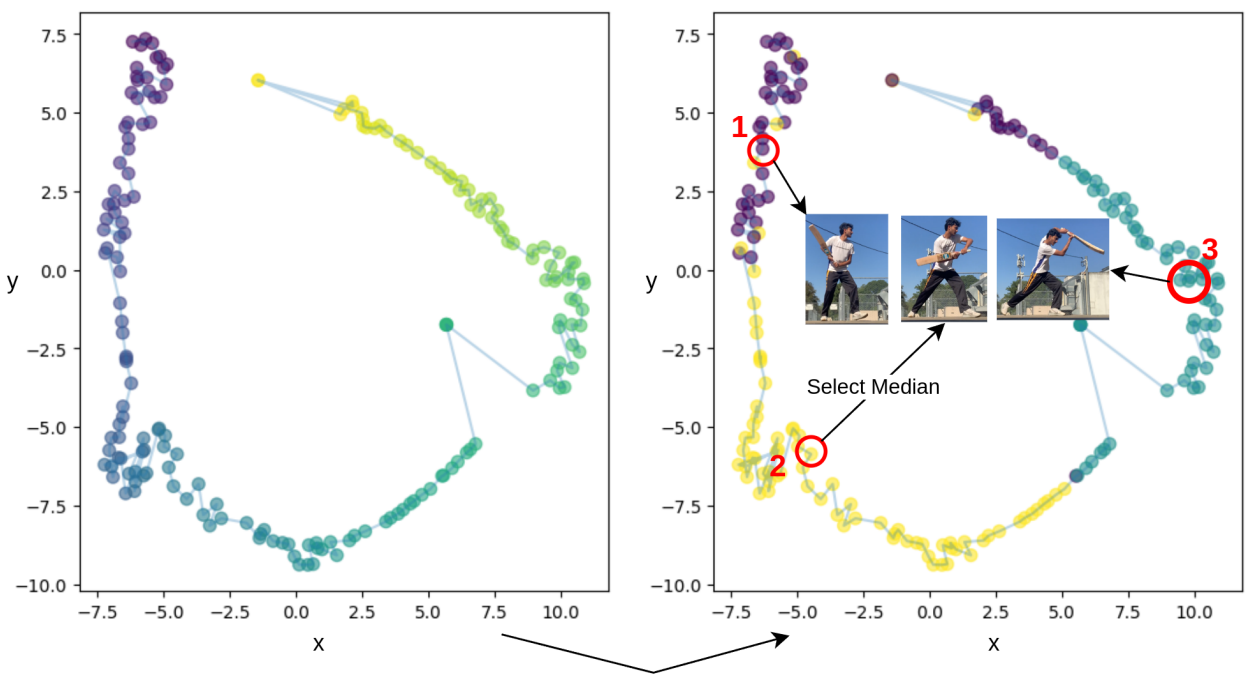

Figure 1: Visualization of motion embeddings and frame selection approaches. Left: 2D t-SNE projection of frame embeddings colored by temporal order, showing trajectory of a cricket shot. Right: Selected key frames using our HMM-based approach that identifies semantically meaningful phases (stance, execution, follow-through) from the continuous motion sequence.

We introduce CricRAG, a retrieval-augmented framework that bridges this gap by explicitly providing VLMs with level-appropriate reference points. Our key insight is that by retrieving similar but slightly better-executed techniques as benchmarks, we can guide VLMs to generate feedback that suggests achievable improvements rather than unrealistic comparisons to professional standards. This approach enables a curriculum-based learning experience that better mirrors human coaching practices while leveraging the powerful visual understanding capabilities of VLMs.

Our technical contributions include:

- A novel annotated dataset of 288 cricket technique videos spanning four fundamental shots.
- A retrieval pipeline combining motion embedding and contrastive learning, with 78% top-3 retrieval accuracy.
- A frame sampling optimization using Gaussian Hidden Markov Models that reduces VLM inference costs while maintaining feedback quality (Figure 1)
- Evaluation demonstrating that retrieval-augmented alignment significantly improves VLM coaching feedback

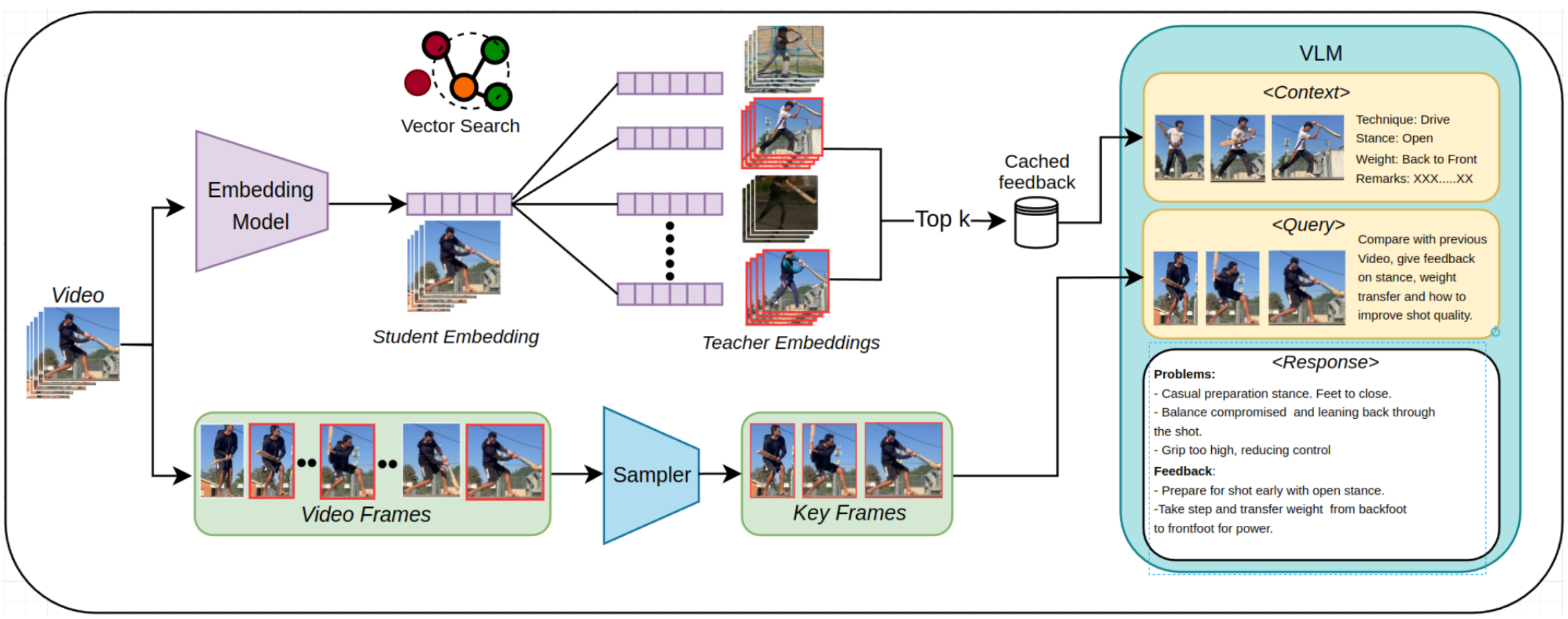

Figure 2: CricRAG's pipeline for personalized cricket coaching: The input video is processed through parallel streams: (1) An embedding model projects frames into a learned technique space for similarity-based retrieval of appropriate reference videos, (2) A frame sampler identifies key technique phases to optimize VLM inference, and (3) The retrieved reference frames and sampled query frames are combined with attribute-specific prompts to generate level-appropriate coaching feedback.

## Related Work

Sports analysis through computer vision has evolved from basic pose estimation (Cao et al. 2017) to sophisticated motion analysis (Zhang, Black, and Tang 2021). In cricket, existing work has focused primarily on professional-level applications like shot classification (Paramanik, Joseph, and Kumar 2024) and detection (Datta et al. 2024). While these advances demonstrate the potential of automated analysis, they typically target high-level performance assessment rather than developmental coaching. Our work specifically addresses the needs of developing players by introducing level-appropriate benchmarking and feedback generation.

Recent advances in VLMs have enabled sophisticated visual understanding and instruction generation (Alayrac et al. 2022). However, their application to sports coaching presents unique challenges, particularly in providing contextually appropriate feedback (Lu et al. 2023). VLMs struggle with in-context learning for visual tasks, often failing to adapt their assessment criteria based on skill level or learning context. CricRAG addresses these limitations through explicit retrieval-based alignment with coaching principles.

Retrieval-Augmented Generation (RAG) has emerged as a powerful approach for enhancing language model performance through external knowledge integration (Lewis et al. 2020). While RAG has been extended to multimodal applications (Zhao et al. 2023), its use in motion analysis and sports coaching remains unexplored.

Traditional approaches to automated sports coaching have often relied on complex sensor setups or professional motion capture systems, limiting their accessibility. Recent work has begun exploring more practical solutions using standard video input, but these systems typically lack the adaptability needed for developmental coaching. CricRAG bridges this gap by combining the accessibility of video-based analysis with curriculum-aware feedback generation through retrieval-augmented VLM alignment.

### CricRAG Dataset

Existing sports datasets often focus on action recognition, judging or pose estimation, lacking the granular feedback crucial for coaching (Parmar and Tran Morris 2019), (Zhou et al. 2023). There is no dataset on cricket technique. Cricket techniques present unique challenges due to the complexity of motion sequences and the need for skill-level appropriate feedback. We create the novel CricRAG dataset. The CricRAG dataset encompasses four fundamental cricket techniques: front-foot defence, back-foot defence, straight drive, and miscellaneous shots. Each video spans 1.5-2.5 seconds, captured at standard smartphone frame rates (25-60 fps), providing sufficient temporal resolution to analyze critical technique components. Professional coaches provided binary labels for three critical attributes: Stance, Weight Transfer, Shot Quality. The CricRAG dataset distribution and details is provided in supplementary materials. The dataset is available upon request via email to the authors.

## Method

CricRAG operates through three components (Figure 2): Motion embedding and retrieval for identifying appropriate benchmark videos, Frame sampling optimization for efficient VLM inference, and VLM alignment through retrieved exemplars for personalized feedback

### Motion Embedding and Retrieval

Retrieving similar cricket techniques requires a robust embedding pipeline that captures both spatial pose information and temporal motion dynamics. Our approach combines three key components: pose estimation, motion encoding, and similarity learning.

First, we extract frame-level pose information using AlphaPose (Fang et al. 2022):

$$\mathbf{p}_t = AlphaPose(\mathbf{I}_t) \quad (1)$$

where $\mathbf{p}_t$ represents the pose keypoints for frame $t$ and $\mathbf{I}_t$ is the input image. AlphaPose provides reliable keypoint detection across varying recording conditions.

Next, we encode the temporal dynamics using MotionBERT (Zhu et al. 2023), which creates view-invariant and body-invariant representations of human motion:

$$\mathbf{e} = MotionBERT(\mathbf{p}_1, ..., \mathbf{p}_T) \quad (2)$$

where $\mathbf{e} \in R^{8704}$ captures the complete motion sequence. MotionBERT's pretrained representations are particularly effective at capturing the subtle weight transfers and balance shifts crucial for cricket technique analysis.

Finally, we learn technique-specific embeddings through a MLP projection head which transforms the motion features into a compact 512-dimensional space optimized for technique retrieval. We train this projection using supervised contrastive loss:

$$\mathcal{L} = \sum_{i \in I} \frac{-1}{|P(i)|} \sum_{p \in P(i)} \log \frac{\exp(\mathbf{z}_i \cdot \mathbf{z}p/\tau)}{\sum a \in A(i) \exp(\mathbf{z}_i \cdot \mathbf{z}_a/\tau)} \quad (3)$$

where $\mathbf{z}_i$ is the projection of the anchor video $i$'s motion embedding, $\mathbf{z}_p$ is the projection of a positive example from set $P(i)$ containing same-technique videos, $\mathbf{z}_a$ is the projection of videos from set $A(i)$ containing all other videos, and temperature $\tau$ controls embedding separation. $|P(i)|$ denotes the number of positive examples for anchor $i$. This loss encourages videos of the same technique to cluster while pushing different techniques apart, enabling effective retrieval of similar but slightly better executions for coaching feedback.

## Frame Sampling

Frame sampling optimization is crucial for practical deployment of smartphone-captured cricket technique analysis. Three key challenges motivate this need: (1) VLM inference costs scale linearly with frame count ($2.70-4.00$ per 1000 images), (2) Vision transformers struggle with long sequences(Kim et al. 2024), leading to diluted attention across similar frames, and (3) Most VLMs have strict input limitations on images per query. Through empirical validation with cricket coaches, we confirmed that 3-5 well-chosen frames capture the essential information for technique assessment, corresponding to the three key phases in cricket batting: stance (initial positioning), execution (bat-ball contact), and follow-through (shot completion). We employ a Gaussian Hidden Markov Model trained using the Expectation Maximization algorithm (implemented via hmmlearn) to identify these phases from the frame embeddings, which are first reduced to 50 dimensions using PCA. The model discovers natural groupings in the motion sequence through unsupervised learning, and we select representative frames from each identified phase using median frame selection to ensure temporal coherence. This approach ensures robust frame selection while avoiding boundary transitions where phase assignments might be ambiguous.

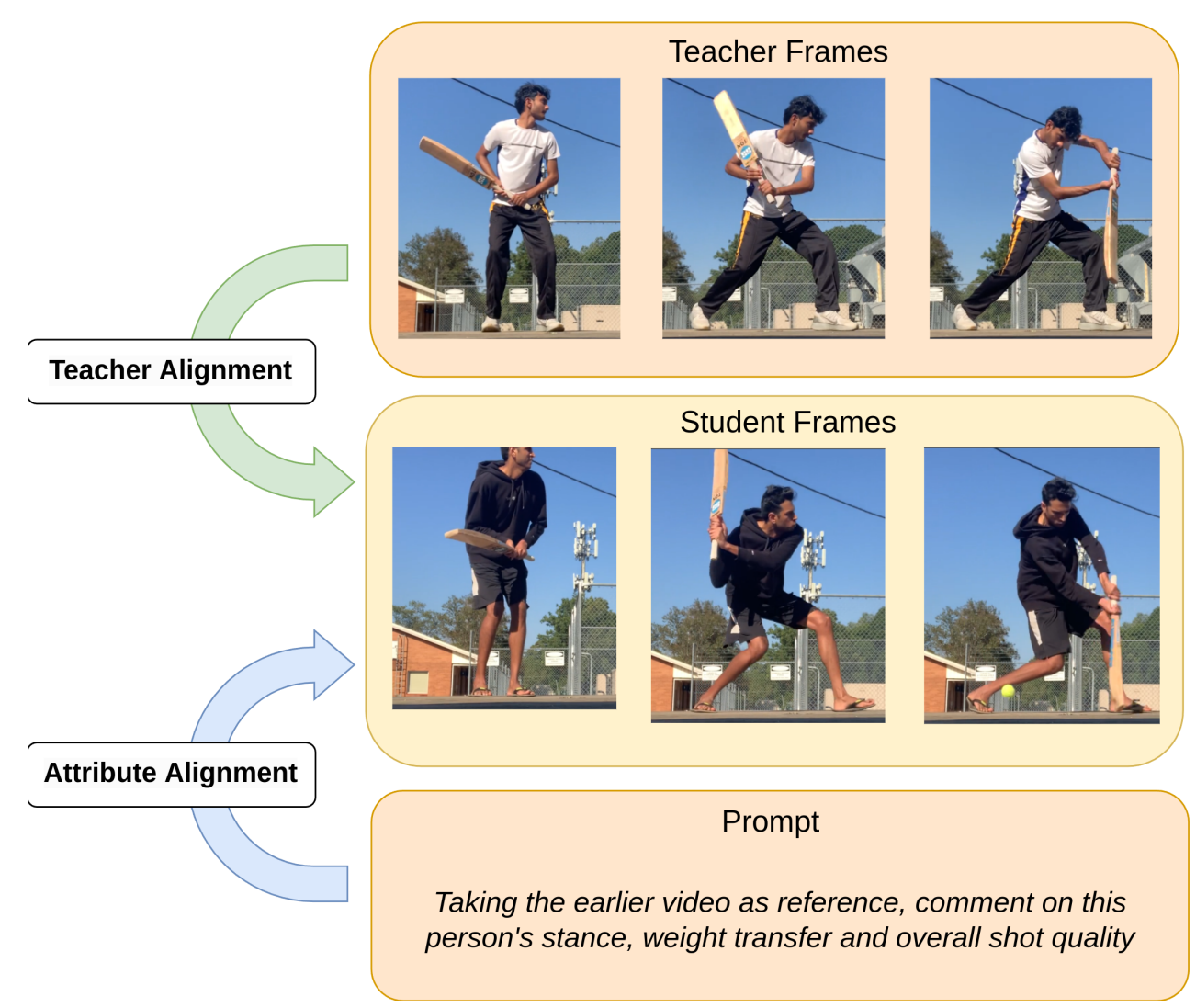

Figure 3: Demonstration of VLM alignment through teacher-student frame comparison. Top: Retrieved reference (”teacher”) frames showing proper technique execution. Middle: Student frames being evaluated. Bottom: Structured prompting approach that guides the VLM to focus on specific technical attributes (stance, weight transfer, shot quality) while using the reference frames as an appropriate benchmark.

## VLM Alignment Through Retrieval

We formulate VLM alignment as a guided feedback task where $F = VLM(Context = I_{r1}, I_{r2}, I_{r3}, Query = I_{q1}, I_{q2}, I_{q3}, A)$, where $I_{ri}$ are frames from the retrieved reference, $I_{qi}$ are query frames, $A$ specifies attributes to evaluate, and $F$ is the generated feedback. The frames contribute to teacher alignment and benchmarking, and the text prompt aligns the VLM towards the attributes we want to analyze, as shown in Figure 3.

# Results

## Retrieval Performance

Table 1 shows retrieval accuracy across MLP configurations. The single-layer 512-dimensional configuration significantly outperforms alternatives, achieving 65% top-1 and 78% top-3 accuracy.

| Hidden Dim | Projection Dim | Top-1 | Top-3 |
|---|---|---|---|
| 2048, | 1024 | 0.61 | 0.67 |
| 1024 | 256 | 0.63 | 0.69 |
| 1024 | 512 | **0.65** | **0.78** |

Table 1: Retrieval accuracy across MLP configurations.

While our top-k accuracy provides a useful proxy for retrieval performance, it has limitations for cricket techniques that exist on a continuum rather than as discrete classes. A shot combining elements of multiple techniques (e.g., a drive with defensive characteristics) could retrieve a semantically similar shot labeled with a different class, leading to artificial penalization under this metric. Figure 4 shows the projected embeddings, notice how drive shots (green) maintain proximity to both defensive techniques, reflecting how real cricket shots often combine multiple elements. We provide further discussion and results on other metrics in the supplementary materials. This reflects a broader challenge in evaluating retrieval systems for continuous action spaces.

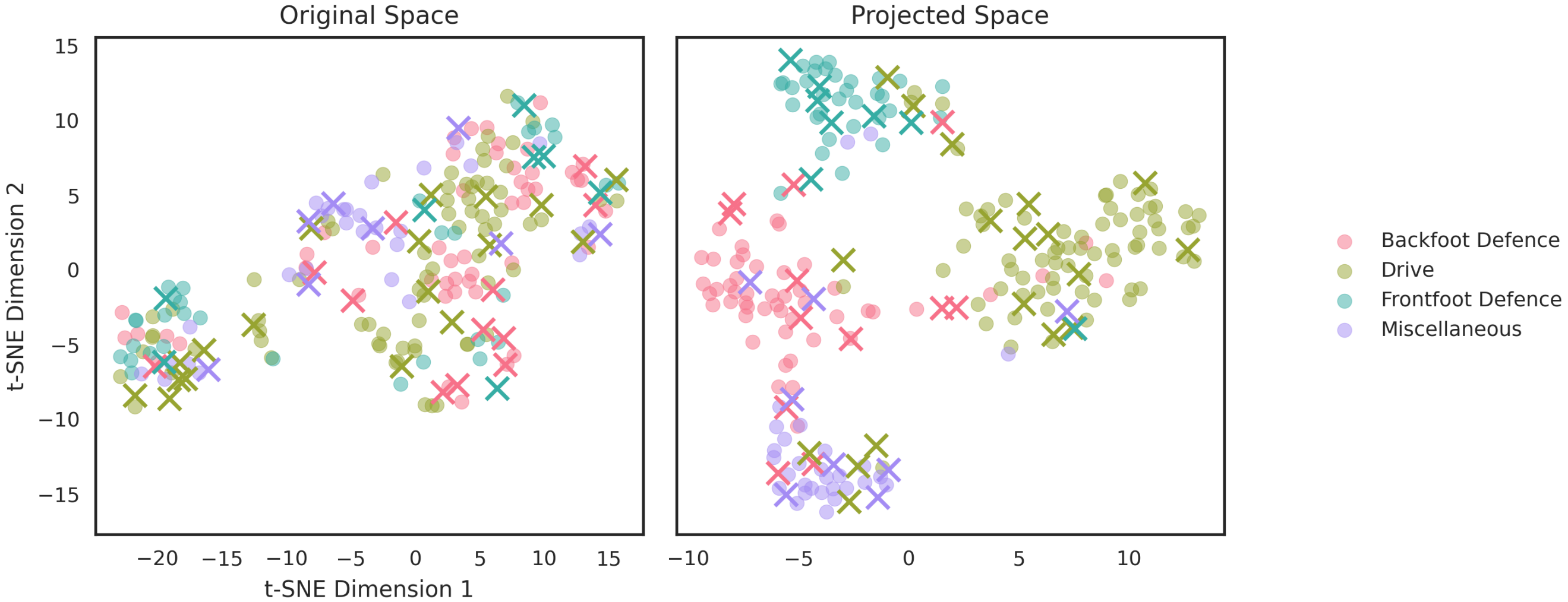

Figure 4: t-SNE visualization comparing motion embeddings before (left) and after (right) contrastive learning with our MLP projection head. The projected space demonstrates improved clustering while preserving the continuous nature of cricket techniques. Training samples (dots) and test samples (X) show how techniques naturally blend.

### Frame Sampling Analysis

Our HMM-based frame selection approach effectively identifies key technical phases in cricket shots (Figure 1). Through empirical testing, we found that using 3 hidden states naturally aligns with the stance, execution, and follow-through phases of cricket techniques. The PCA reduction to 50 dimensions balances computational efficiency with preservation of motion characteristics. While our initial experiments suggest the effectiveness of this semantic approach over time-based sampling methods, comprehensive quantitative evaluation comparing different sampling strategies remains an important direction for future work.

### VLM Performance

We use a 80/20 split between for the RAG dataset (used for retrieval) and test set. For each test video, we use feedback from both, VLM and human coaches on our three key attributes (stance, weight transfer, and shot quality).

To quantify alignment between VLM and human feedback, we employ a separate LLM-as-a-judge using the framework in (Kim et al. 2024). The judge analyzes the semantic similarity between VLM and human feedback for each attribute and classifies wether they are in agreement. We measure accuracy as the proportion of cases where there is agreement between VLM and human feedback, averaged across all attributes and techniques.

| Model | No Context | With Context | $\Delta$ |
|---|---|---|---|
| Qwen2.5-7B | 52% | 68% | +16% |
| Claude Sonnet 3.5 | 67% | **94%** | **+27%** |
| GPT-4V | 71% | 88% | +17% |

Table 2: VLM feedback accuracy with and without retrieval context.

Table 2 shows results from this evaluation process. The dramatic improvement in feedback alignment with retrieval context (+27% for Claude Sonnet 3.5) demonstrates the effectiveness of our approach in guiding VLMs toward human-like coaching assessments. We provide the prompt structure and discussion on performance on individual models in the supplementary materials.

## Conclusion and Future Work

This paper introduces CricRAG, a retrieval-augmented framework that aligns Vision-Language Models with human coaching practices for personalized cricket instruction. We demonstrate that our approach significantly improves coaching feedback quality across multiple VLMs, with Claude 3.5 achieving 94% agreement with human assessments compared to 67% without retrieval context. Several directions remain for future work. Extending to dynamic sports like tennis will require multi-view analysis. Transitioning from text to voice-based interaction could improve real-world usability. These limitations notwithstanding, CricRAG demonstrates how retrieval-augmented approaches can effectively bridge the gap between AI capabilities and human teaching practices.

# Supplementary Materials

## CricRAG Dataset

The CricRAG dataset consists of 288 cricket technique videos spanning four fundamental cricket shots, with comprehensive annotations from professional coaches. The dataset is available upon request via email to the authors.

Each video in the dataset:

- Spans 1.5-2.5 seconds at standard smartphone frame rates (25-60 fps)
- Features a single cricket shot execution
- Is annotated by experienced cricket experts for three key attributes, Stance, Weight transfer effectiveness, and Overall shot quality.
- Includes binary labels (good/bad) for each attribute

Table 3 shows the distribution of cricket techniques in our dataset. The higher proportion of drive shots reflects their fundamental importance in batting technique development. Table 4 presents the attribute-wise quality distribution across techniques, revealing interesting patterns in technique execution and assessment.

Table 3: Dataset composition showing the distribution of cricket techniques. Our dataset contains a balanced mix of fundamental cricket shots, with drives being the most common due to their importance in batting technique.

| Technique | Count | Percentage |
|---|---|---|
| Drive | 112 | 38.9% |
| Backfoot Defence | 77 | 26.7% |
| Frontfoot Defence | 53 | 18.4% |
| Miscellaneous | 46 | 16.0% |
| Total | 288 | 100% |

Table 4: Attribute distribution across cricket techniques. For each technique class, we show the percentage of videos rated as good for each attribute. Stance shows the most variation across techniques, suggesting it is a key discriminating factor in technique assessment.

| 2*Technique | Attributes (% Good) | | |
|---|---|---|---|
| | Stance | Weight Transfer | Shot Quality |
| Drive | 55.4% | 87.5% | 58.4% |
| Backfoot Defence | 53.2% | 88.3% | 64.8% |
| Frontfoot Defence | 62.3% | 81.1% | 66.7% |
| Miscellaneous | 34.8% | 81.4% | 52.5% |

Figure 5 shows representative frames from our dataset, illustrating the variety of techniques and execution styles captured. These frames demonstrate the three key phases of cricket shots that our frame sampling approach aims to identify: stance, execution, and follow-through.

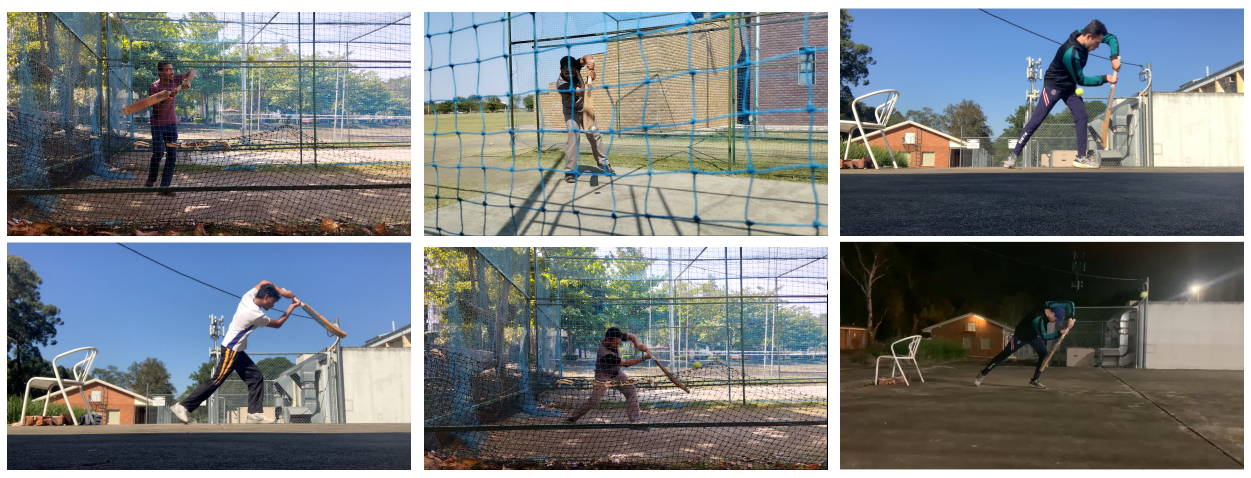

Figure 5: Representative frames from our cricket technique dataset. Each video in our dataset is annotated with stance quality, weight transfer effectiveness, and overall shot quality rating.

## Embedding Space Analysis

To validate the effectiveness of our MLP projection head, we analyze the average distances of technique embeddings to their respective cluster centers before and after projection:

Table 5: Average distances to cluster centers in original vs projected embedding spaces. The reduction in distances after projection indicates successful clustering of similar techniques while maintaining semantic relationships.

| Technique | Original Space | Projected Space |
|---|---|---|
| Back-foot Defence | 9.245 | 5.095 |
| Drive | 9.410 | 5.103 |
| Front-foot Defence | 14.186 | 3.772 |
| Miscellaneous | 8.542 | 5.843 |

The reduction in cluster distances across all techniques demonstrates the projection head's ability to create a more organized embedding space. Notably, front-foot defence shows the most significant improvement (from 14.186 to 3.772), suggesting particularly effective clustering for this technique. The relatively consistent projected distances for back-foot defence and drives (5.095 and 5.103 respectively) aligns with the intuition that these techniques often share similar motion characteristics. The slightly higher distance for miscellaneous shots (5.843) is expected given the greater variance in this category. This analysis complements the retrieval accuracy metrics presented in the main paper by providing direct evidence of improved embedding organization.

## VLM Evaluation Details

We evaluated three Vision-Language Models (VLMs) on cricket technique assessment: Qwen-VL-7B, Claude 3.5 Sonnet, and GPT-4V. Each model was tested on their ability to assess cricket shots with and without reference videos. Here we provide some qualitative observations from the experiments.

**Qwen-VL-7B:**

- Configuration: bfloat16, max pixels: 720×1080, 1.0 fps
- Achieved approximately 50% accuracy on single video assessment
- Consistently rated comparison videos as worse than reference videos, even when passed identical videos.

**Claude 3.5 Sonnet:**

- Without context: Good videos often rated average
- With context: Successfully used reference videos as grounding standard

**GPT-4V:**

- Good assessment of good technique videos without benchmarking
- Less accurate on poor technique videos without reference
- Significantly improved assessment accuracy when given reference frames

## Prompt Structures

We used two types of prompts in our evaluation: standalone assessment and reference-based comparison. Below we detail the structure of each:

**Standalone Assessment** For evaluating a single video without reference:

```
PROMPT = {
  role: "user",
  content: [
    {text: "Here are given three frames
    of a person hitting a cricket shot.
    Comment on their stance,  weight
     transfer and overall shot quality."},
    {image: frame1},
    {image: frame2},
    {image: frame3}
  ]
}
```

**Reference-Based Assessment** For comparing technique against a reference video:

```
PROMPT = {
  role: "user",
  content: [
    {text: "Here are three frames of a
     person hitting a cricket shot."},
    {image: ref_frame1},
    {image: ref_frame2},
    {image: ref_frame3},
    {text: "Here are three more frames.
     Taking the earlier video as reference,
     comment on this person's stance,
     weight transfer and overall shot
     quality."},
    // Query frames
    {image: query_frame1},
    {image: query_frame2},
    {image: query_frame3}
  ]
}
```